\pdfoutput=1
\documentclass[11pt]{article}

\usepackage{emnlp2021}

\usepackage{times}
\usepackage{latexsym}
\usepackage{hyperref}
\usepackage{subfigure}
\usepackage{graphicx}
\usepackage{multirow}
\usepackage{booktabs}
\usepackage{amsmath}
\usepackage{amsthm}
\usepackage{amssymb}
\usepackage{diagbox}
\usepackage{algorithm}
\usepackage{algorithmic}
\usepackage{makecell}
\usepackage{color}
\usepackage[marginal]{footmisc}

\usepackage[T1]{fontenc}
\usepackage[utf8]{inputenc}

\usepackage{microtype}
\definecolor{aqua}{rgb}{0.0, 1.0, 1.0}
\definecolor{arylideyellow}{rgb}{0.91, 0.84, 0.42}
\definecolor{columbiablue}{rgb}{0.61, 0.87, 1.0}
\title{A Three-Stage Learning Framework for Low-Resource Knowledge-Grounded Dialogue Generation}
\author{
Shilei Liu, Xiaofeng Zhao, Bochao Li,  Feiliang Ren$^{*}$, 
Longhui Zhang, Shujuan Yin \\
School of Computer Science and Engineering\\ Key Laboratory of Medical Image Computing of Ministry of Education\\Northeastern University, Shenyang, 110169, China \\
\texttt{liusl@live.cn}, 
\texttt{renfeiliang@cse.neu.edu.cn} \\
}
\begin{document}
\maketitle
\footnote{$^*$ Corresponding author.}
\begin{abstract}
Neural conversation models have shown great potentials towards generating fluent and informative responses by introducing external background knowledge. Nevertheless, it is laborious to construct such knowledge-grounded dialogues, and existing models usually perform poorly when transfer to new domains with limited training samples. Therefore, building a knowledge-grounded dialogue system under the low-resource setting is a still crucial issue. In this paper, we propose a novel three-stage learning framework  based on weakly supervised learning which benefits from large scale ungrounded dialogues and unstructured knowledge base. To better cooperate with this framework, we devise a variant of Transformer with decoupled decoder which facilitates the disentangled learning of response generation and knowledge incorporation. Evaluation results on two benchmarks indicate that our approach can outperform other state-of-the-art methods with less training data, and even in zero-resource scenario, our approach still performs well.
\end{abstract}

\section{Introduction}
Neural dialogue systems have made rapid progress in recent years thanks to the advances in sequence generation technology \citep{DBLP:journals/corr/VinyalsL15,vaswani_attention_2017}. Though such models in neural architectures are able to reply with plausible responses regarding to dialogue history, people can still feel a clear gap when they converse with the chatbots, compared with the conversation with humans. 
To bridge the gap and generate fluent and informative responses, a number of approaches have been proposed by leveraging external knowledge. Knowledge-grounded dialogue is a task of generating an informative response based on both dialogue history and a collection of external knowledge \citep{dinan_wizard_2019}. 
The forms of knowledge are diverse, and in this work, we only focus on knowledge in the form of unstructured documents.

Generally, it is difficult to construct large scale conversations that are naturally grounded on the documents for learning of a response generation model \citep{zhao_low-resource_2020}, and most of the previous methods  \citep{lian_learning_2019,DBLP:conf/acl/LiNMFLZ19,kim_sequential_2020,dinan_wizard_2019} perform poorly when transfer into a new domain with limited training samples. So there are growing appeals for low-resource dialogue response generation, which aims to leverage past experience to improve the performance with limited labeled training examples of target corpus.

\begin{figure*}[htbp]
  \centering
  \includegraphics[width=16cm]{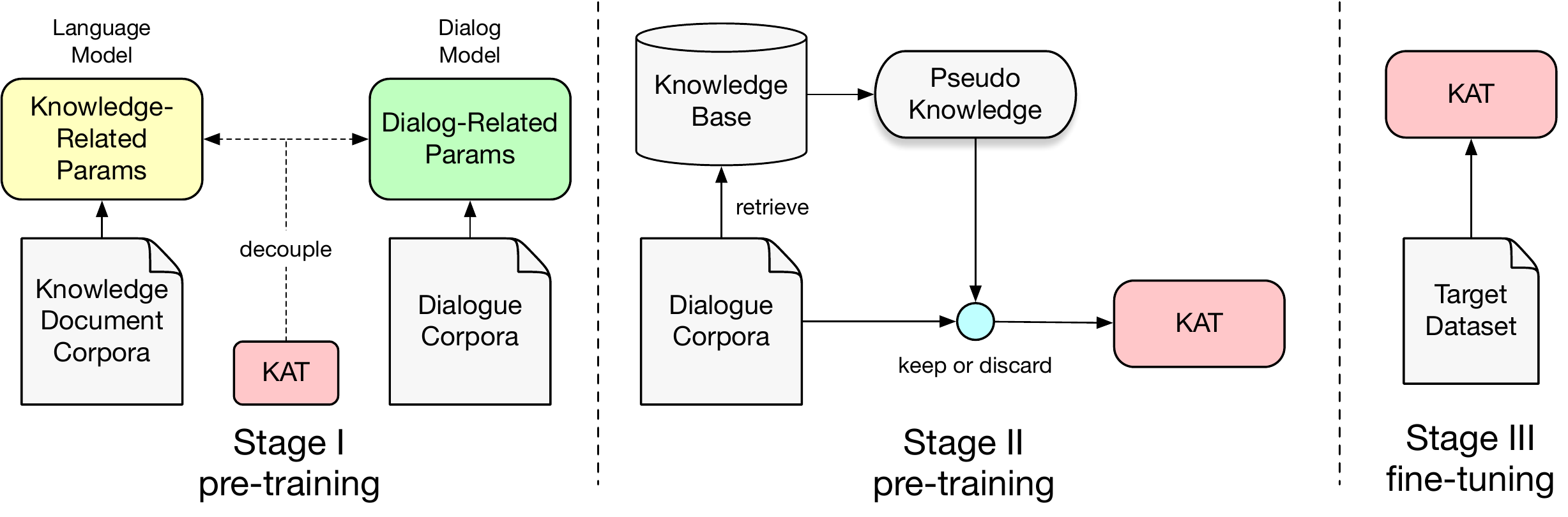}
  \caption{Our three-stage learning framework (TSLF).}
  \label{fig:stages}
\end{figure*}

To address this issue, we envisage to absorb useful information from other easily accessible heterogeneous datasets to enhance the performance of the knowledge-based dialogue model under low-resource setting. 
Based on this assumption, we propose a novel \textbf{Three-Stage Learning Framework (TSLF)}. TSLF attempts to divide the parameters of a model into dialogue-related and knowledge integration-related. 
In the first stage, we use supervised learning to pre-train dialogue-related parameters on general dialogues (e.g., online forum comments), and perform domain-adaptive pre-training \citep{DBLP:conf/acl/GururanganMSLBD20} to initialize knowledge-related parameters on unlabeled knowledge base (e.g., items in Wikipedia). 
In the second stage, inspired by the distant supervision in the relation extraction \citep{DBLP:conf/acl/MintzBSJ09}, we match a set of pseudo-knowledge for each ungrounded dialogue to construct a lower quality knowledge-grounded dialogue dataset, and further co-pretrain the above two groups of parameters on this dataset. 
In the third stage, the trained model will be fine-tuned on the target low-resource dataset. The flow of TSLF is shown in Figure \ref{fig:stages}.

\begin{figure}[htbp]
  \centering
  \includegraphics[width=7cm]{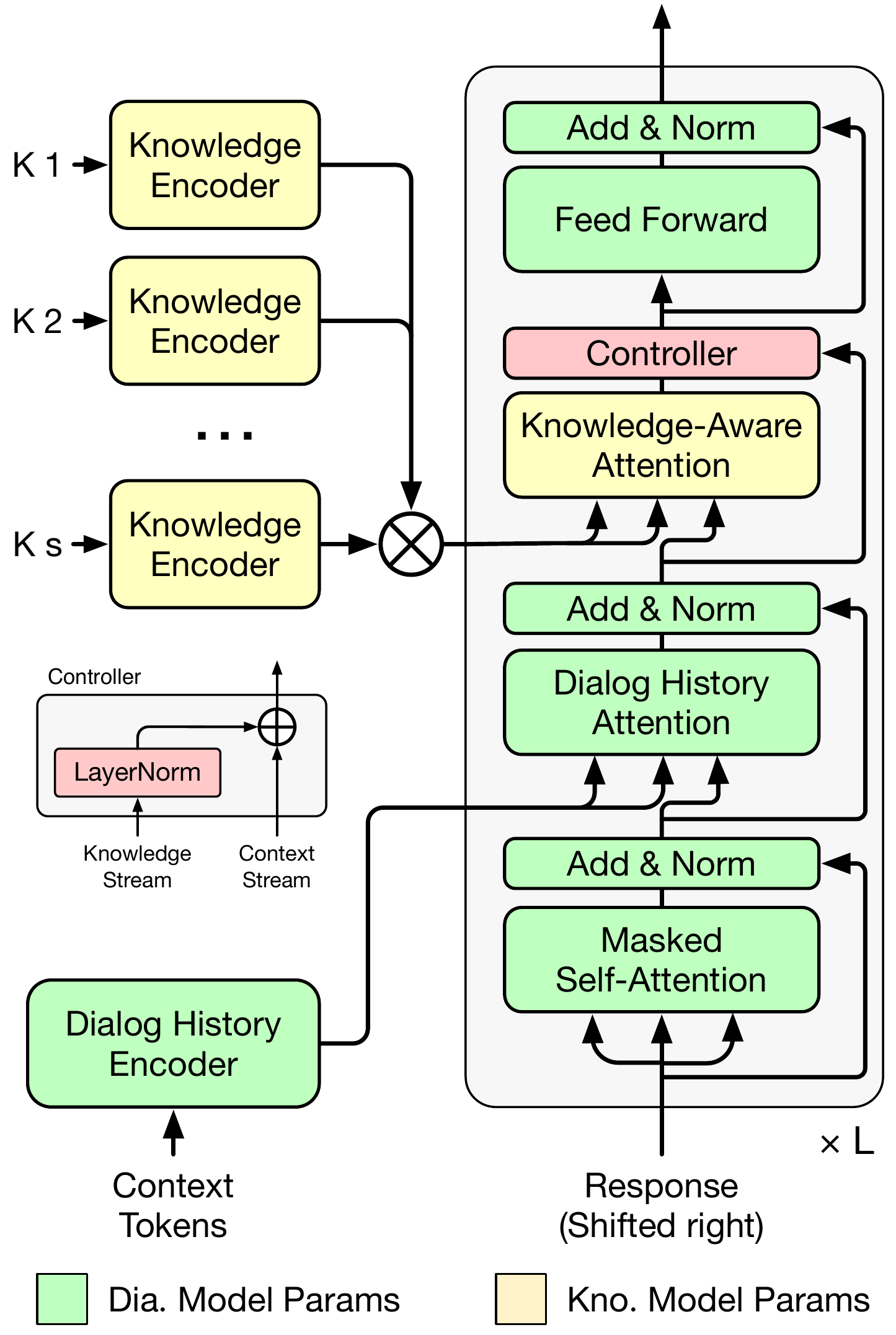}
  \caption{The architecture of our KAT.}
  \label{fig:kat}
\end{figure}

In order to better cooperate with the disentangled learning mechanism in TSLF, we devise \textbf{Knowledge-Aware Transformer (KAT)}, a variant of vanilla Transformer \citep{vaswani_attention_2017} whose parameters are decoupled that facilitates the separate learning of dialogue generation and knowledge incorporation. As shown in Figure \ref{fig:kat}, besides dialogue history, KAT also accepts a set of knowledge as additional input. KAT has a knowledge-aware decoder which could obtains information from the dialogue context and background documents through cross-attention and integrates them through a controller. 

We conduct experiments on two knowledge-grounded dialogue generation benchmarks including Wizard-of-Wikipedia \citep{dinan_wizard_2019} and CMU\_DoG \citep{DBLP:conf/emnlp/ZhouPB18}. Evaluation results in terms of both automatic metrics and human judgment indicate that using only about 1/4 of the training data on Wizard (1/16 on CMU\_DoG), the performance of our approach outperforms the competitive baselines which are learned from full crowd-sourced training corpora. Even without using any training data of the target dataset, our method still performs well.

The contributions in this work are summarized as follows: 
(1) We propose a novel three-stage learning framework that leverages weakly supervised learning to help build a low-resource knowledge-grounded dialogue generation model; 
(2) We devise knowledge-aware Transformer, a knowledge-grounded neural conversation model with a novel dynamic knowledge selection mechanism, which can fully exploits the external knowledge to generate fluent and informative dialogue responses; 
(3) Our KAT-TSLF achieves surprising performance under the scenarios of full data, low-resource and even zero-resource. 

The source code is available at \url{https://github.com/neukg/KAT-TSLF}. 

\section{Approach}
Low-resource knowledge-grounded dialogue generation is task that requires a method to learn from experience $E$, which consists of direct experience $E_d$ containing limited monolingual context-knowledge-response triples and indirect experience $E_i$, to improve the performance in response generation measured by the evaluation metric $P$. The direct experience $E_d$ refers to the training samples of target corpus $\mathcal{D}_l=\{(U_i, \mathcal{K}_i, Y_i)\}_{i=1}^{m_1}$ ($U_i$ is dialog history, $Y_i$ is response, and    $\mathcal{K}_i=\{K_j\}_{j=1}^{s}$ is a set of external knowledge documents of $i$-th sample) which are under low-resource settings. In this work, we consider $E_i$ as a large scale ungrounded dialogue dataset $\mathcal{D}_d=\{(U_i, Y_i)\}_{i=1}^{m_2}$, a knowledge base $\mathcal{D}_k=\{K_i\}_{i=1}^{m_3}$ ($m_2, m_3 \gg m_1$) and a pre-trained language model which are easy to obtain. In the following, we first introduce our KAT, and then show how to train it from coarse to fine under our TSLF. 

\subsection{Knowledge-Aware Transformer}
KAT accepts $U$ and $\mathcal{K}=\left\{K_i\right\}^s_{i=1}$ as inputs, and generates a response $\hat{Y}$.
It consists of three components: a dialogue context encoder (DE) to encode $U$, a knowledge encoder (KE) to encode $\mathcal{K}$, and a decoder to incorporate dialog history, dynamically select knowledge and generate response. 
The architecture of KAT is shown in Figure \ref{fig:kat}.

\subsubsection{Encoder}
We define DE as a Transformer encoder, and the output is represented as $\mathbf{U} \in \mathbb{R}^{n \times d}$, where $n$ is the sequence length, and $d$ is the hidden state dimension. Similarly, KE is defined as another Transformer encoder, and it encode each document individually. Following KE is a concatenation operation that concatenates all document representations: $\mathbf{K} = \left[\mathbf{K}_1; ...; \mathbf{K}_s\right] \in \mathbb{R}^{sz  \times d}$, where $\mathbf{K}_i \in \mathbb{R}^{z  \times d}$ is output of $i$-th KE, and $z$ is the sequence length of each document. $\mathbf{K}$ and $\mathbf{U}$ will be used for the input of the decoder.

\subsubsection{Knowledge-Aware Decoder}
Generally, not all knowledge in the $\mathcal{K}$ contributes to the generation of the response,
so the model should have the ability to select knowledge.
Different from \citep{dinan_wizard_2019,lian_learning_2019,kim_sequential_2020} who perform knowledge selection in the encoding phase (or in a pipeline), we leaves it to the decoding phase.
Based on the Transformer decoder, we propose a cross attention based decoder which can
select knowledge dynamically and generate informative response. 

\paragraph{Knowledge Integration Block (KIB)}
As shown in the right part of Figure \ref{fig:kat}, we add a new block after the dialogue history attention block in Transformer decoder layer. It takes the output from last block as \emph{query}, and the memory from $\mathbf{K}$ as \emph{key} and \emph{value}. The output of this block can be obtained by

multi-head attention mechanism \citep{vaswani_attention_2017}. During decoding, KIB can dynamically select different knowledge according to dialogue context and the tokens that have been generated at current time step.

\paragraph{Controller}
To control the knowledge and context contributions in each layer, we add a gate after the knowledge selection block.
Denote $\mathbf{h}_k$ as output of KIB and $\mathbf{h}_c$ as the residual from the previous block, the output of controller can be expressed by
\begin{equation}
  \begin{split}
    &\text{CT}(\mathbf{h}_k,\mathbf{h}_c) = \beta \cdot \text{LN}(\mathbf{h}_k) + (1-\beta)\cdot \mathbf{h}_c \\
    &\beta = \sigma \left(\mathbf{w}\cdot \left[\mathbf{h}_k; \mathbf{h}_c\right]\right)
  \end{split}
\end{equation}
where $\mathbf{w} \in \mathbb{R}^{2d}$ is a learnable parameter and $\sigma$ denotes sigmoid function.

\subsection{Three-Stage Learning Framework}\label{sec:3stages}
For further discussion, we denote $\theta_d$, $\theta_k$, and $\theta_a$ as the learnable parameters of the green, yellow and pink parts in Figure \ref{fig:kat} respectively.   We can observe that $\theta_d$ is related to context encoding and response generation, $\theta_k$ is related to knowledge representation and integration, and these two parts are disentangled. 
In order to benefit from a wealth of heterogeneous corpora, we propose a three-stage learning framework.  
In TSLF, we first initialize $\theta_d$ and $\theta_k$ in a decoupled scheme by training in ungrounded dialogues and unstructured knowledge documents respectively, and then co-optimize them with $\theta_a$ by weakly supervised learning and finally transfer KAT to target low-resource dataset. The illustration of TSLF is shown in Figure \ref{fig:stages}.
\subsubsection{Stage I}
We choose the state-of-the-art Transformer based encoder-decoder model BART \citep{lewis_bart_2020} as the the backbone, pre-training it on $\mathcal{D}_d$ with dialogue response generation task:
\begin{equation}
  \mathcal{L}_{d}(\theta_d)=-\sum_{(U,Y)\in \mathcal{D}_d}\sum_{t}\log p(y_t|y_{<t},U)
\end{equation}

Besides, inspired by \citet{DBLP:conf/acl/GururanganMSLBD20}, we conduct domain-adaptive pre-training on unlabeled knowledge documents to improve knowledge representation ability.
Specifically, 15\% of tokens in a text $K$ are replaced with \texttt{<mask>} or noise  words, and another Transformer tries to rebuild it:
\begin{equation}
  %\small
  \mathcal{L}_{k}(\theta_k^+)=-\sum_{K\in \mathcal{D}_k}\sum_{t}\log p(k_t|k_{<t},\hat{K})
\end{equation}
where $\hat{K}$ is the corrupt $K$. We disentangle the encoder and the cross-attention block in each decoder layer from this Transformer ($\theta_k^+$) and initialize $\theta_k$ with them.  
\subsubsection{Stage II} \label{sec:stage2}
In stage I, $\theta_d$ and $\theta_k$ are trained separately, and the connection between knowledge and dialogue has not yet been established. If KAT is fine-tuned directly on low-resource dataset $\mathcal{D}_k$, it may cause inconsistency problems, so we add a warm-up process to it.

Intuitively, responses from humans carry clues to relevance of the knowledge candidates \citep{zhao_knowledge-grounded_2020}, so the knowledge document that promotes the flow of dialogue usually has a high textual similarity with the response. Based on this assumption, we construct a set of pseudo-knowledge for some dialogues in $\mathcal{D}_d$ to form a new weak supervision dataset $\mathcal{D}_p$ according to Algorithm \ref{alg:build}.  

$\mathcal{I}(query, documents)$ means retrieve the document with the highest similarity (e.g., TF-IDF and BM25). Context-response pairs with low quality will be removed. In the knowledge-grounded dialogue corpora, only less documents in knowledge pool are valuable, and others are noise. The design of negative samples is to simulate this situation and make the distribution of knowledge in $\mathcal{D}_p$ closer to the target data set.

We perform weakly supervised learning on $\mathcal{D}_p$ to warmup KAT:
\begin{equation}
  \mathcal{L}(\theta_d,\theta_k,\theta_a)=-\sum_{(U,\mathcal{K},Y)\in \mathcal{D}_p}\log p(Y|\mathcal{K},U)
\end{equation}
\begin{algorithm}[tb]
\caption{Construction of $\mathcal{D}_p$}
\label{alg:build}
\textbf{Input}: Ungrounded dialogues $\mathcal{D}_d$, documents $\mathcal{D}_k$, threshold $\gamma$ and number of negative samples $o$;\\
\textbf{Output}: $\mathcal{D}_p$;
\begin{algorithmic}[1] %[1] enables line numbers
\STATE Initialize $\mathcal{D}_p=\phi$;
\FOR{$(U, Y)$ \textbf{in} $\mathcal{D}_d$}
\STATE $K$, $score$ = $\mathcal{I}(Y, \mathcal{D}_k)$;
\IF {$score > \gamma$}
\STATE $\mathcal{K}=\{K\}$;
\FOR{$i$ \textbf{in} $\{1,...,o\}$}
\STATE Sample $K^{\prime}$ from $\mathcal{D}_k - \mathcal{K}$ randomly;
\STATE $\mathcal{K} \gets \mathcal{K} \cap \{K^{\prime}\}$;
\ENDFOR
\STATE $\mathcal{D}_p \gets \mathcal{D}_p \cap \{(U, \mathcal{K},Y)\}$;
\ENDIF
\ENDFOR
\STATE \textbf{return} $\mathcal{D}_p$;
\end{algorithmic}
\end{algorithm}

\subsubsection{Stage III}
After warming up on $\mathcal{D}_p$, KAT will be fine-tuned on the target low-resource dataset: % with a small batch size:
\begin{equation}
  \mathcal{L}(\theta_d,\theta_k,\theta_a)=-\sum_{(U,\mathcal{K},Y)\in \mathcal{D}_l}\log p(Y|\mathcal{K},U)
\end{equation}

If not fine-tuned, KAT can also be directly applied to zero-resource response generation. 
\begin{table*}[htbp]
  \centering
  %  \scriptsize
  \small
  \begin{tabular}{lccccccccc}
    %\specialrule{.08em}{0pt}{0pt}
    \toprule
    Models                                          & PPL  & BLEU-1 & BLEU-2 & BLEU-3 & BLEU-4 & R-1  & R-2 & DIST-1 & DIST-2 \\ \midrule
    ITDD \citep{DBLP:conf/acl/LiNMFLZ19}            & 17.8 & 15.8   & 7.1    & 4.0    & 2.5    & 16.2 &  -   &   -     &  -      \\
    $\text{BART}_{cat}$                             &   19.7  & 23.1   & 11.4   & 6.7    & 4.3    & 19.3 & 5.1 & 7.1    & 29.9   \\
    $\text{BART}_{skt}$ \citep{kim_sequential_2020} & 20.3 & 23.2   & 11.9   & 7.6    & 4.4    & 19.4 & 5.4   & 6.8   & 30.3   \\
    DRD \citep{zhao_low-resource_2020}       & 23.0 & 21.8   & 11.5   & 7.5    & 5.5    & 18.0 & -   & -      & -      \\
    ZRKGC$^\dag$ \citep{li_zero-resource_2020}     & 40.4 & 22.2   & 7.3    & 2.8    & 1.8    & 18.6 & 2.4 & 5.4    & 22.5   \\ \midrule
	KAT Full Data                                       & 14.5  & 25.5       & 13.9    & 9.0    & 6.6   & 21.6    & 7.5 &  9.3   & 37.0      \\
    KAT-TSLF Full Data                                       & 14.4     & 25.5     &  13.9    &  9.1 &  6.7   &  21.7  &7.6  & 9.5   &  38.3       \\
    KAT-TSLF 1/4 Data                                        &  17.6    &  23.3    &   12.2     &  7.7      &   5.5     &  20.3    & 6.8    &    9.9    &  39.1      \\
    KAT-TSLF 1/8 Data                                       &  18.8    &  22.5      &  11.5      &  7.1      &   4.9     &  19.8    & 6.3    &   9.9     &  39.5      \\
    KAT-TSLF Zero Data                                       & 100+    & 19.5   & 8.1    &  4.0   & 2.2    & 14.7  & 3.0  & 7.5   & 33.9   \\ \bottomrule
  \end{tabular}
  \caption{Evaluation results on Wizard test seen. $\dag$ marks zero-resource setting. The results of ITDD and DRD are copied from \citep{zhao_low-resource_2020} and DRD is under full-data. The performance of \emph{KAT-TSLF 1/4 Data} outperforms $\text{BART}_{cat}$ and $\text{BART}_{skt}$ significantly except BLEU-1 (t-test with $p$-value < 0.01, the same table below).}
  \label{tab:wowseen}
\end{table*}
\begin{table*}[htbp]
  \centering
  \small
  \begin{tabular}{lccccccccc}
    %\specialrule{.08em}{0pt}{0pt}
    \toprule
    Models                & PPL  & BLEU-1 & BLEU-2 & BLEU-3 & BLEU-4 & ROUGE-1 & ROUGE-2 & DIST-1 & DIST-2 \\ \midrule
    ITDD                  & 44.8 & 13.4   & 4.7    & 2.1    & 1.1    & 11.4    &  -      &  -     & -      \\
    $\text{BART}_{cat}$ & 24.5  & 23.2   & 11.0   & 6.3    & 4.1    & 18.9    & 4.5     & 5.3    & 22.2   \\
    $\text{BART}_{skt}$   & 22.3 & 23.4   & 10.9   & 6.8    & 4.6    & 19.0    & 4.7    & 5.2   & 24.5   \\
    DRD                   & 25.6 & 20.7   & 10.1   & 6.2    & 4.3    & 16.5    & -       & -      & -      \\
    ZRKGC$^\dag$                 & 41.5 & 21.8   & 7.1    & 2.7    & 1.1    & 18.5    & 2.4     & 3.4    & 15.6   \\ \midrule
    KAT                                        & 15.8  & 24.4       & 12.5    & 7.8    & 6.6   & 20.5    & 6.4 &  10.1   & 39.1      \\
    Full Data             &  15.8  &  24.1      &12.9&  8.3   &  6.0    & 20.7 &    7.2     &   6.7     & 26.0       \\
    1/4 Data             &  18.4    &  23.1    &   11.9     &   7.5     &   5.2     &  19.9       &     6.4    &   6.6     &  25.1      \\
    1/8 Data             & 20.1     & 22.3    &    11.3    &  7.0   &  4.8   &  19.0   &    5.9     &      6.6  &    25.3     \\
    Zero Data             & 100+    & 19.6   & 8.6    &  4.7   & 2.7    & 14.9    &  3.0    & 5.7   &  26.4   \\ \bottomrule
  \end{tabular}
  \caption{Evaluation results on Wizard-of-Wikipedia test unseen.}
  \label{tab:wowunseen}
\end{table*}
\section{Experiments}
\subsection{Datasets and Evaluation Methods}
We conduct extensive experiments on two public English knowledge-grounded datasets: Wizard-of-Wikipedia \citep{dinan_wizard_2019} and CMU\_DoG \citep{DBLP:conf/emnlp/ZhouPB18}. Wizard-of-Wikipedia is a chit-chatting dataset between two agents, and the two participants are not quite symmetric: one will play the role of a knowledgeable expert (which we refer to as the wizard) while the other is a curious learner (the apprentice). Each wizard turn is associated with $\sim$60 sentences retrieved from the Wikipedia and each sentence contains $\sim$30 words, and most of them are noise. The test set is split into two subsets, test seen and test unseen. The difference between the two is that the former contains some topics that overlap with the training set.  CMU\_DoG also contains conversations between two workers who know the background documents and try to discuss the content in depth. Different from Wizard-of-Wikipedia which spans multiple topics, CMU\_DoG mainly focuses on film reviews. 

Reddit Conversation Corpus is a large scale open domain dialogue corpus cleaned by \citet{DBLP:journals/corr/abs-1811-01063} which consists of $\sim$15M samples for training and $\sim$0.8M samples for validation. Following \citet{zhao_low-resource_2020,li_zero-resource_2020}, we merge the training and validation data of RedditCC as $\mathcal{D}_d$. Besides, we split $\sim$0.5M Wikipedia articles provided by ParlAI\citep{DBLP:conf/emnlp/MillerFBBFLPW17} into $\sim$6.6M sentences as $\mathcal{D}_k$. Information retrieval function $\mathcal{I}$ mentioned in Sec. \ref{sec:stage2} is implemented by Apache Lucene with BM25 algorithm and the size of $\mathcal{D}_p$ is $\sim$0.1M. $\gamma$ and $o$ are set to 16.4 and 39 respectively.

Following the common practice in evaluating open domain dialogue generation, we choose perplexity (PPL), corpus-level BLEU  \citep{DBLP:conf/acl/PapineniRWZ02}, sentence-level ROUGE \citep{linrouge} and corpus-level DISTINCT \citep{DBLP:conf/naacl/LiGBGD16} as metrics. Response with higher BLEU and ROUGE is closer to the ground-truth, and response with higher DIST  has a larger vocabulary that could express more information. BLEU is computed with NLTK library \citep{DBLP:conf/acl/Bird06} and ROUGE is calculated with the code published with \citet{kim_sequential_2020}. 

Besides quantitative evaluation, we also recruit three human annotators to do qualitative analysis on response quality. For each dataset, we randomly sample 100 samples, and each sample contains the conversation history, response, and external knowledge set (for Wizard-of-Wikipedia, we only provide ground-truth knowledge). The annotators then judge the quality of the responses from three aspects, including context coherence, language fluency and knowledge relevance, and assign a score in \{0, 1, 2\} to each response for each aspect. Each response receives 3 scores per aspect, and the agreement among the annotators is measured via Fleiss’ kappa \citep{fkappa}.
\begin{table*}[htbp]
  \centering
  \small
  \begin{tabular}{lccccccccc}
    %\specialrule{.08em}{0pt}{0pt}
    \toprule
    Models                & PPL  & BLEU-1 & BLEU-2 & BLEU-3 & BLEU-4 & ROUGE-1 & ROUGE-2 & DIST-1 & DIST-2 \\ \midrule
    ITDD                  & 26.0 & 9.5   & 3.6     & 1.7    & 0.9    & 10.4 & -         &  -  &  -      \\
    $\text{BART}_{cat}$ &  36.4  & 17.0   & 8.6   & 5.3    & 3.4    & 13.6    & 3.1     & 1.5    & 7.3   \\
    $\text{BART}_{skt}$   & 40.1 & 16.2   & 8.3   & 5.1    & 3.1    & 12.7    & 2.6    & 1.2   & 7.3   \\ 
    DRD                   & 54.4 & 15.0   & 5.7    & 2.5    & 1.2    & 10.7    & -       & -      & -      \\
    ZRKGC$^\dag$                 & 53.5 & 15.1   & 4.2    & 1.2    & 0.4     & 12.5    & 0.7        &  1.2   & 8.1    \\ \midrule
    KAT         &22.2    & 19.4  &  10.5  &  6.9   &  4.7   & 14.4     &  3.3     &   1.8     &  8.9     \\
    Full Data             &  21.7    &  20.4   &  10.6  &  6.7  & 4.4    & 15.1      &   3.7   &  2.0      &   11.1     \\
    1/8 Data             &  25.7    &    19.1    &  10.1      & 6.5    &   4.4     &    13.9     &   3.2      &  1.9      &  10.5      \\
    1/16 Data             &  28.1    &    18.5    &  9.8      & 6.3    &   4.2     &    13.4     &   2.9      &  1.8      &  9.9      \\
    Zero Data             &   100+  &  12.8   &  4.7   &  2.4    &  1.4    &   7.9   &   1.0   & 2.6 & 15.7     \\ \bottomrule
  \end{tabular}
  \caption{Evaluation results on CMU\_DoG. The performance of \emph{KAT-TSLF 1/16 Data} outperforms $\text{BART}_{cat}$ and $\text{BART}_{skt}$ significantly except ROUGE-1 and ROUGE-2 (t-test with $p$-value < 0.01).}
  \label{tab:cmudog}
\end{table*}
\begin{figure*}[htbp]
  \centering
  \includegraphics[width=16cm]{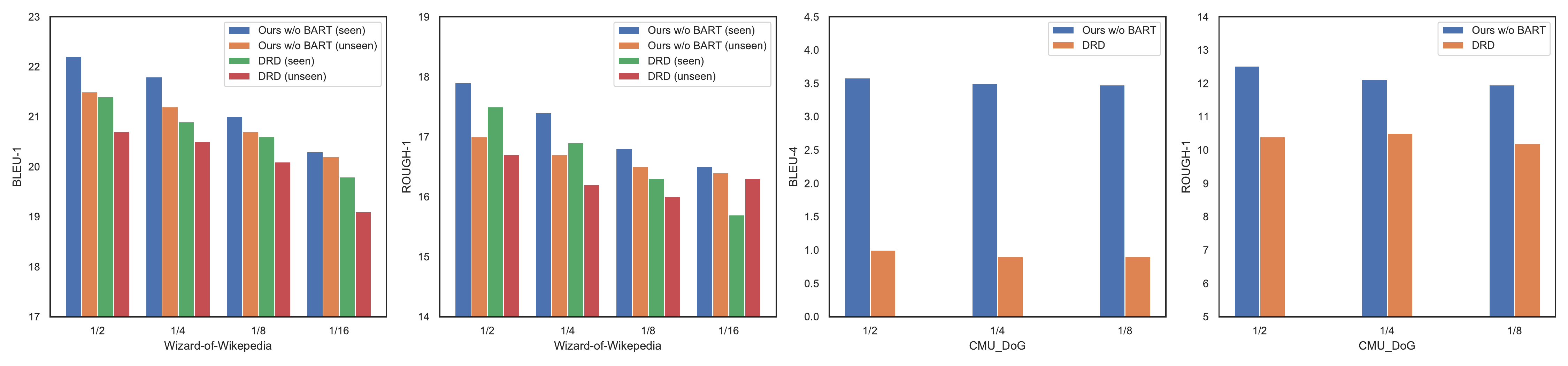}
  \caption{Comparison with DRD in low-resource setting. DRD does not provide results when the training data is less than 1/16 (1/8 in CMU\_DoG). In order to save space, we merge the Wizard seen and unseen into one subfigure.}
  \label{fig:lr}
\end{figure*}
\subsection{Baselines}
We compare our approach with the following baselines:
(1) \textbf{ITDD}: an Transformer-based architecture which incrementally represents multi-turn dialogues and knowledge, and conducts response decoding in two passes \citep{DBLP:conf/acl/LiNMFLZ19};
(2) \textbf{$\text{BART}_{cat}$}: A simple BART-based model that take the concatenation of dialogue context and all knowledge as the input of BART for response generation. BART sets constraint on the maximum number of tokens it can handle, and we directly truncate the text that exceeds the length limit;
(2) \textbf{$\text{BART}_{skt}$}: SKT is variational model that introduced BERT on the basis of  \citet{lian_learning_2019} and considered the knowledge selection history in multi-turn dialogue \citep{kim_sequential_2020}. We feed the knowledge candidate selected by SKT to BART for response generation. It should be noted that training SKT requires human labels that indicate ground-truth knowledge which are crucial to the performance of the model. For fair comparison, we use $\mathcal{I}$ to reselect the knowledge label;
(3) \textbf{DRD}: Another low-resource dialogue model which devise a disentangled response decoder with copy mechanism \citep{see_get_2017} and use a two-stage framework to learn it \citep{zhao_low-resource_2020}. DRD is not open source, so we can't make a very detailed comparison with it;
(4) \textbf{ZRKGC}: A double latent variable model that achieves the state-of-the-art performance in zero-resource knowledge-grounded dialogue generation \citep{li_zero-resource_2020}. ZRKGC is based on UNILM \citep{DBLP:conf/nips/00040WWLWGZH19} with 110M parameters whose performance is close to BART, so we will not replace the backbone of ZRKGC.
%For fair comparison, baselines (1) - (4) are pre-trained using the RedditCC dataset.
\subsection{Implementation Details}
The knowledge pool of target dataset is usually very large (e.g. $\sim$60 sentences in Wizard), in order to reduce the time overhead, following \citep{kim_sequential_2020}, we only keep the first 40 sentences. 
We use the base version of BART with 139M parameters in our work, and the number of parameters of KAT is 196M.
The batch size in stage I, II and III is 2048, 128 and 16 respectively. 
The max sequence length in source and target is 256 and 64 respectively. 
All models are optimized with AdamW \citep{DBLP:journals/corr/abs-1711-05101} with learning rate $5e-5$ in 3 epochs. 
We employ beam search in response decoding (the number of beams from 1 to 3) implemented by \citet{DBLP:conf/emnlp/WolfDSCDMCRLFDS20}. 
\begin{table*}[htbp]
  \centering
  \small
  \begin{tabular}{lccccccccccc}
    \toprule
    \multirow{2}{*}{Models} & \multicolumn{4}{c}{Wizard Test Seen} & \multicolumn{4}{c}{Wizard Test Unseen} & \multicolumn{3}{c}{CMU\_DoG}                                                      \\ \cmidrule(lr){2-5}\cmidrule(lr){6-9}\cmidrule(lr){10-12}
                            & CC                                   & LF                                     & KR                           & Kappa & CC   & LF   & KR & Kappa & CC & LF & Kappa \\ \midrule
    $\text{BART}_{skt}$     & 1.78                                 & 1.80                                   & 1.34                         & 0.61  & 1.72  &1.74 & 1.36 & 0.64   & 1.70 &1.72 & 0.65   \\
    ZRKGC                   & 1.72                                 & 1.75                                   & 1.12                         & 0.63  & 1.69 & 1.70 & 1.16&  0.63  &  1.67 &1.69& 0.63  \\
    Ours 1/8 Data           & 1.81                                 & 1.82                                   & 1.35                         & 0.63  & 1.79 & 1.78  & 1.35 & 0.66  & 1.74 & 1.75 & 0.69   \\
    Ours Zero Data          & 1.76                                 & 1.78                                   & 1.14                         & 0.64  & 1.70 & 1.72 & 1.24 &  0.64  & 1.69 & 1.71 & 0.66      \\\bottomrule
  \end{tabular}
  \caption{Human evaluation results on Wizard-of-Wikipedia and CMU\_DoG. CC, LF and KR marks
    \emph{context coherence}, \emph{language fluency} and \emph{knowledge relevance} respectively. In zero-resource setting, our KAT-TSLF outperforms ZRKGC. Besides, our model surpasses $\text{BART}_{skt}$ (full data) in most metrics with only only 1/8 of the training data.}
  \label{tab:humaneval}
\end{table*}
\subsection{Evaluation Results}
\begin{figure*}[htbp]
  \centering
  \includegraphics[width=16cm]{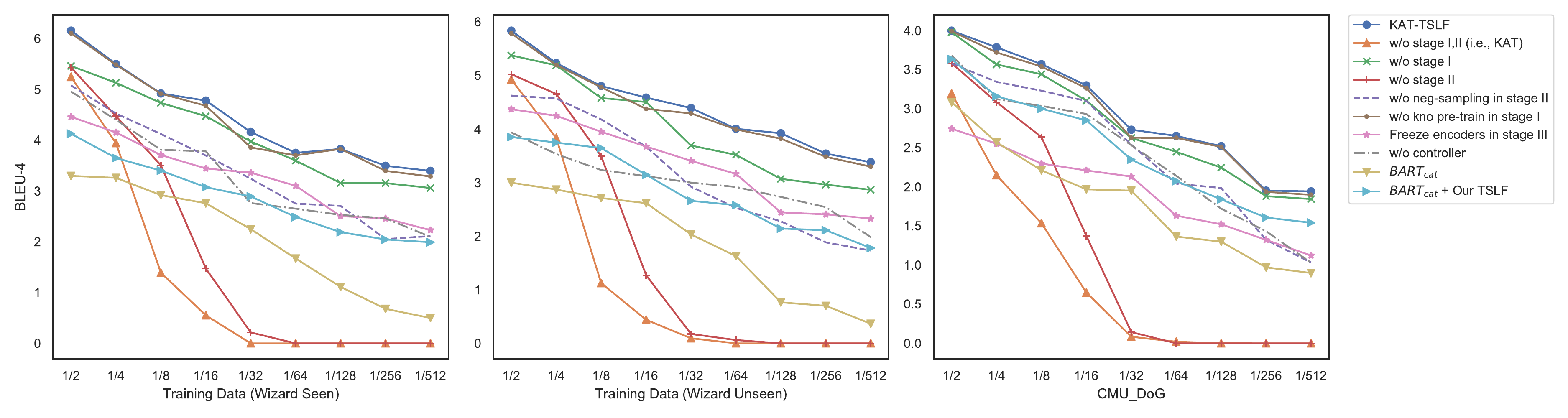}
  \caption{Ablation experiments on Wizard-of-Wikipedia and CMU\_DoG. The number of beams are set to 1 for all models. It is recommended to view the picture after zooming in, and the more the curve is to the upper right, the better the result.}
  \label{fig:abl}
\end{figure*}

Table \ref{tab:wowseen}, \ref{tab:wowunseen} and \ref{tab:cmudog} reports the evaluation results on automatic metrics, and we have the following observations: 
(1) In the full-data scenario, KAT achieves state-of-the-art performance without using any additional corpora, which means that KAT itself is an excellent dialogue model. 
Besides, additional resources are unnecessary when there are enriched training datas, so TSLF has little effect in this setting; 
(2) KAT-TSLF achieves the comparable performance with $\text{BART}_{cat/skt}$ even though the baselines have leveraged all training data, while our model is only learned with 1/4 training data on Wizard (1/16 on CMU\_DoG). 
We compare the low-resource performance with DRD, and the results are shown in Figure \ref{fig:lr}. 
For a fair comparison, we removed the pre-training language model and reduce the number of model parameters. 
We can see that KAT-TSLF outperforms DRD (especially in CMU\_DoG). The comparison with $\text{BART}_{cat}$ is supplemented in Figure \ref{fig:abl}; 
(3) Although our TSLF is mainly for low-resource scenarios, under the setting of zero resources (i.e., without stage III), the performance of KAT-TSLF also surpasses ZRKGC in most evaluation metrics; 
(4) Responses generated by KAT have higher DIST-$n$, which means that our KAT can better obtain information from multiple knowledge and generate more diverse texts.

Table \ref{tab:humaneval} reports the human evaluation results. We observe that responses from our KAT-TSLF are more fluent and more contextually coherent than those from $\text{BART}_{skt}$ and ZRKGC. Compared with our low-resource model, SKT has stronger knowledge relevance in the case of full data, thanks to its well-designed knowledge selection module.

\subsection{Ablation Study}
We conduct ablation experiments on Wizard and CMU\_DoG, and the results are shown in Figure \ref{fig:abl}. 

So as to verify the effect of TSLF, we first removed stage I, stage II, and stage I II respectively. 
Inserting a new module into an already well-trained large-scale pre-trained language model will cause inconsistency problems, which require a lot of data to reconcile, so after removing stage II or stage I II, the performance of our KAT in low-resource dropped sharply. 
Although the quality of the automatically constructed warm-up dataset $\mathcal{D}_p$ is lower than the target dataset $\mathcal{D}_l$, it also helps to establish the connection between the knowledge representation component and the dialogue component. 
Besides, we tried not to pre-train $\theta_k$ on unlabeled documents, and the result has dropped slightly, which demonstrates that is still helpful to tailor a pre-trained model to the domain of a target task. 
In addition, replacing negative sampling with top-k retrieval will increase the inconsistency with the  knowledge distribution of target dataset, leading to performance degradation. 
Moreover, the controller also has an effect on the generalization of the model. It can help KAT quickly adapt to new domains by adjusting the proportion of knowledge and context in the response. In order to improve the generalization performance with limited training data, some works \citep{DBLP:journals/corr/abs-2102-09397,zhao_low-resource_2020} fix most of the parameters during fine-tuning. We also tried to frozen knowledge encoder and context encoder in stage III or stage II III, and the results show that the performance has not improved, indicating that with the help of stage II, our model can hardly  fall into overfitting.

In order to verify the effect of our TSLF on other models, we try to combine $\text{BART}_{cat}$ with TSLF. Since the parameters of BART are tightly coupled, we can only apply stage II to it. Experimental results show that the performance is improved significantly under low-resource setting.
\subsection{Discussions}
\paragraph{Case Study} Table \ref{tab:case1} shows a case from Wizard, from which we can see that the response from our model with zero data not only smoothly catches the ground-truth knowledge (highlighted in blue), but also expands the topic with proper pieces of other knowledge (highlighted in yellow). ZRKGC generated sentences that were inconsistent with the facts.  Although $\text{BART}_{skt}$ chose the correct knowledge, the narrative was too straightforward, and there is a repetition phenomenon. 
We showed some other cases in the supplementary material. 
\paragraph{Comparison with DRD} If we ignore the details, DRD is actually a special case of our method, which skips stage II. During pre-training, DRD completely separates dialogue-related components and knowledge representation-related components, which makes it difficult to effectively promote the integration of dialogue and knowledge with only a small number of samples during fine-tuning. So when the training data is extremely small, DRD can hardly work. 
Besides, in order to prevent overfitting, DRD has to limit the number of parameters of the knowledge integration component and use fix other parameters when fine-tuning, which leads to limited performance of the model. 
In addition, the complex model structure makes it difficult for DRD to use pre-trained language models.
\paragraph{KAT v.s. $\text{BART}_{cat}$} BART (as well as most other pre-training language models) has a limit on the maximum tokens of the input, so useful knowledge is likely to be truncated. For example, there are about 60 external documents per sample in Wizard, and about 40 documents will be truncated. In theory, KAT can accept an unlimited number of knowledge, so this should be one of the reasons why KAT's performance is better than $\text{BAER}_{cat}$. When we reduce the maximum number of knowledge that KAT can handle (a hyperparameter) to 15, the performance is close to $\text{BART}_{cat}$. 
\begin{table}[htbp]
\small
\centering
\begin{tabular}{ll}
\toprule
\begin{tabular}[c]{@{}l@{}}Dial.\\Hist.\end{tabular}    & \begin{tabular}[c]{@{}l@{}}A: Yea it was a great movie. The Last of the \\ Mohicans was released in 1992. \\B: I didn't realize it's been out that long! \\What is it about?\end{tabular}
        \\ \midrule
\begin{tabular}[c]{@{}l@{}}GT \\Kno.\end{tabular}       & \begin{tabular}[c]{@{}l@{}}The Last of the Mohicans is a 1992 American \\epic historical drama, set in 1757 during the \\French and Indian War.\end{tabular}
        \\ \midrule
Ref.    & \begin{tabular}[c]{@{}l@{}}Well The Last of the Mohicans is an epic\\ historical drama. It was set in 1757 during \\the Indian and French war.\end{tabular}           \\ \midrule
\multicolumn{2}{l}{\begin{tabular}[c]{@{}l@{}}
\textbf{($\text{BART}_{skt}$)} It's about the French and Indian War. It's \\about the French and Indian War. \\
\textbf{(ZRKGC)} It 's a classic movie. The Last of My Moh-\\icans was released in 2016, and is still out on Netflix.         \\
\textbf{(Ours Zero Data)} It's a series of short stories set \colorbox{columbiablue}{in} \\\colorbox{columbiablue}{1757 during the French and Indian War} \colorbox{arylideyellow}{in the Adi-}\\ \colorbox{arylideyellow}{rondack mountains of Virginia}.     \\ 
\textbf{(Ours 1/16 Data)} It is about a group of people who \\ fight to keep their independence from the French and \\Indian War.       

\end{tabular}}                                                                                                                                            \\ \bottomrule
\end{tabular}
\caption{A case from test seen of Wizard-of-Wikipedia. This dialogue contains a total of 40 external knowledge, one of which is marked as ground-truth (GT).}
\label{tab:case1}
\end{table}

\section{Related Work}
Open domain end-to-end dialogue response generation is inspired by the success of applying neural sequence to sequence models on machine translation \citep{DBLP:conf/nips/SutskeverVL14,DBLP:journals/corr/BahdanauCB14,vaswani_attention_2017}.
Very recently, in order to generate fluent, coherent and informative response, many approaches have been proposed by introducing external background documents \citep{ghazvininejad_knowledge-grounded_2018,yavuz_deepcopy_2019,DBLP:conf/acl/LiNMFLZ19,lin_generating_2020}. 
Besides documents \citep{dinan_wizard_2019,DBLP:conf/emnlp/ZhouPB18}, the are many forms of knowledge such as images \citep{DBLP:conf/chi/HuberMBGD18}  and triples in knowledge graph \citep{DBLP:conf/acl/WuGZWZLW19,DBLP:conf/emnlp/TuanCL19}.  

\citet{dinan_wizard_2019} presents to divide knowledge-grounded dialogue into two steps: knowledge selection and dialogue generation. PostKS \citep{lian_learning_2019}, SKT \citep{kim_sequential_2020}, PIPM \citep{chen_bridging_2020} and SKT-KG \citep{zhan_augmenting_2021} use the prior and posterior distribution of knowledge to improve the accuracy of knowledge selection. \citet{zhao_knowledge-grounded_2020} devise a reinforcement learning method to train a knowledge selector without ground-truth knowledge label. DeepCopy \citep{yavuz_deepcopy_2019}, ITDD \citep{DBLP:conf/acl/LiNMFLZ19} and KIC \citep{lin_generating_2020} have improved the structure of the decoder so that it can better integrate knowledge.
Since knowledge-guided dialogue corpora need to be constructed through crowdsourcing, the size of datasets  such as Wizard-of-Wikipedia \citep{dinan_wizard_2019} are relatively small.
\citet{zhao_low-resource_2020} and \citet{li_zero-resource_2020} proposed to conduct the knowledge-grounded conversation under the low-resource and zero-resource settings respectively. 
We do not compare with \citet{lin_generating_2020,zhao_knowledge-grounded_2020}  since they did not release their entire source codes.

Our three-stage learning framework is inspired by \citet{zhao_low-resource_2020}, which uses ungrounded dialogues and unstructured documents to train a knowledge-grounded dialogue model that can work in low-resource situations. 
In addition, the design of stage II is inspired by distant supervision technology in relation extraction task \citep{DBLP:conf/acl/MintzBSJ09}. 
The idea of KAT is also encouraged by disentangled decoder \citep{DBLP:conf/naacl/RaghuGM19} and the recent breakthrough in variants of Transformer \citep{DBLP:conf/acl/LiNMFLZ19,DBLP:conf/sigir/HashemiZC20,DBLP:journals/corr/abs-2007-01282}. 

\section{Conclusion}
We study knowledge-grounded dialogue generation under a low-resource setting by proposing a three-stage learning framework and a knowledge-aware Transformer. Evaluation results on two benchmarks indicate that our model achieves the state-of-the-art performance with less training data. 
Besides, KAT-TSLF exhibits a good generalization ability on zero-resource scenario. 

\section*{Acknowledgments}
This work is supported by the National Natural Science Foundation of China (No.U1708261 and No. 61572120), Shenyang Medical Imaging Processing Engineering Technology Research Center (17-134-8-00), the Fundamental Research Funds for the Central Universities (No. N181602013 and No.N2016006), Ten Thousand Talent Program (No.ZX20200035), and Liaoning Distinguished Professor (No.XLYC1902057).
\section*{Broader Impact}
Incorporating knowledge into dialogue systems has been the pursuit of researchers in this field for many years. This kind of system will make AI dialogue more natural definitely. It will be more favored by people when the technology does not require a large amount of artificially annotated data. More importantly, the knowledge-based dialogue system can fundamentally change the experience of human-machine dialogue, because system can develop with the update of external knowledge base. One day it will be true that people can obtain effective information through simple conversations. However, coins always have two sides. In addition to the well-known problems caused by large pre-trained datasets for end-to-end dialogue models, special knowledge bases which may be deliberately tailored can also be used to make the generated dialogues biased, just as search engines inadvertently spread biased content created by someone. In order to prevent this technology from being abused, we look forward to more research effort for detecting fake/biased/offensive content. At the same time, we recommend that developers choose content carefully to build a knowledge base for the dialogue system. Good external knowledge can adjust the behavior of the dialogue model in the response process and help the model overcome the biases hidden in large-scale social media datasets.

\bibliography{anthology,custom}
\bibliographystyle{acl_natbib}

%\appendix
%
%\section{Example Appendix}
%\label{sec:appendix}
%
%This is an appendix.

\end{document}